\documentclass[10pt,twocolumn,letterpaper]{article}

\usepackage{cvpr}              
\definecolor{cvprblue}{rgb}{0.21,0.49,0.74}
\usepackage[pagebackref,breaklinks,colorlinks,allcolors=cvprblue]{hyperref}
\usepackage{booktabs}
\usepackage{multirow}
\usepackage[table]{xcolor} 
\usepackage{adjustbox} 
\usepackage[T1]{fontenc}
\usepackage[utf8]{inputenc}
\def\paperID{*****} 
\def\confName{CVPR}
\def\confYear{2026}

\title{%
\makebox[0pt][l]{%
  \hspace*{-1.3em}
  \raisebox{0.45em}{\includegraphics[height=1.8em,trim={0 1mm 0 0},clip]{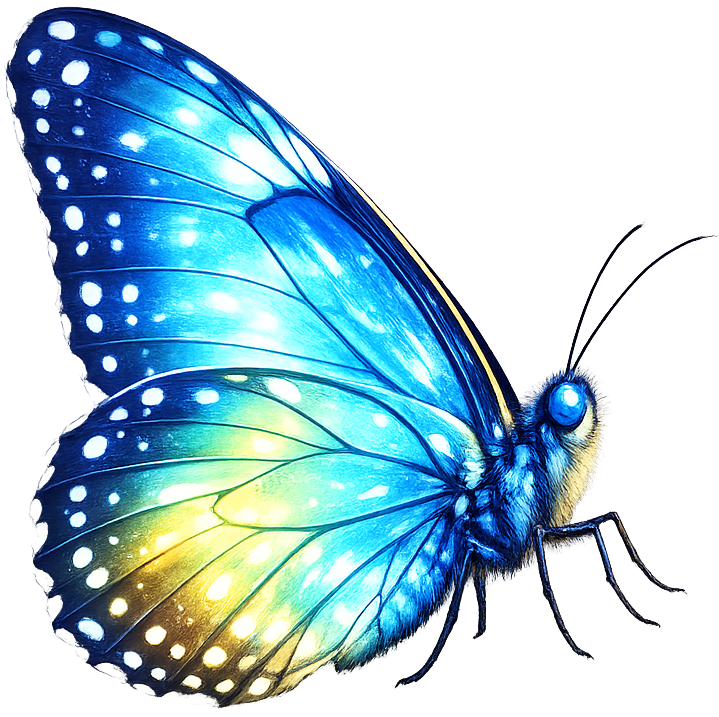}}%
}%
BTS-rPPG: Orthogonal Butterfly Temporal Shifting \\
for Remote Photoplethysmography
}

\author{
{\large
Ba-Thinh Nguyen$^{1}$,
Thi-Duyen Ngo$^{1}$,
Thanh-Trung Huynh$^{3}$,
Thanh-Ha Le$^{1,\dagger}$,
Huy-Hieu Pham$^{2,3,4,\dagger}$
}\\[0.6em]
$^{1}$VNU University of Engineering and Technology, Hanoi, Vietnam\\
$^{2}$VinUni-Illinois Smart Health Center, VinUniversity, Hanoi, Vietnam\\
$^{3}$College of Engineering and Computer Science, VinUniversity, Hanoi, Vietnam\\
$^{4}$Center for Innovations in Health Sciences, VinUniversity, Hanoi, Vietnam\\
[0.4em]
{\tt\small
\{22028163, duyennt, ltha\}@vnu.edu.vn,
\{trung.ht, hieu.ph\}@vinuni.edu.vn
}
}

\begin{document}
\maketitle

\begingroup
\renewcommand{\thefootnote}{\fnsymbol{footnote}}
\setcounter{footnote}{0}
\footnotetext[2]{Corresponding authors: hieu.ph@vinuni.edu.vn, ltha@vnu.edu.vn}

\endgroup
\begin{abstract}
Remote photoplethysmography (rPPG) enables contactless physiological sensing from facial videos by analyzing subtle appearance variations induced by blood circulation. However, modeling the temporal dynamics of these signals remains challenging, as many deep learning methods rely on temporal shifting or convolutional operators that aggregate information primarily from neighboring frames, resulting in predominantly local temporal modeling and limited temporal receptive fields. To address this limitation, we propose BTS-rPPG, a temporal modeling framework based on Orthogonal Butterfly Temporal Shifting (BTS). Inspired by the butterfly communication pattern in the Fast Fourier Transform (FFT), BTS establishes structured frame interactions via an XOR-based butterfly pairing schedule, progressively expanding the temporal receptive field and enabling efficient propagation of information across distant frames. Furthermore, we introduce an orthogonal feature transfer mechanism (OFT) that filters the source feature with respect to the target context before temporal shifting, retaining only the orthogonal component for cross-frame transmission. This reduces redundant feature propagation and encourages complementary temporal interaction. Extensive experiments on multiple benchmark datasets demonstrate that BTS-rPPG improves long-range temporal modeling of physiological dynamics and consistently outperforms existing temporal modeling strategies for rPPG estimation.
\end{abstract}    
\section{Introduction}
\label{sec:intro}

Physiological signals such as heart rate provide valuable insights into human health, emotional state, and physical activity. In clinical practice, these signals are commonly measured using contact sensors such as electrocardiography (ECG) and photoplethysmography (PPG), which require devices to be attached directly to the skin. Although these techniques provide reliable measurements, their contact-based nature limits their applicability in many everyday scenarios where unobtrusive or continuous monitoring is desired.
\begin{figure}[t]
  \centering
  \includegraphics[width=\linewidth]{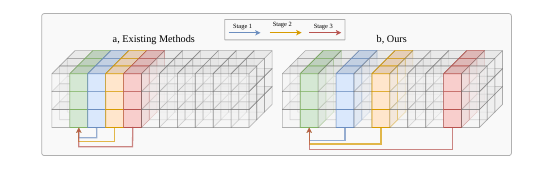}
  \caption{Comparison of temporal communication patterns between (a) existing methods and (b) the proposed BTS-rPPG. Existing methods mainly exchange features between adjacent frames, limiting long-range temporal interaction. In contrast, BTS-rPPG employs a butterfly-inspired hierarchical communication scheme, analogous to FFT-style sparse multi-stage mixing, to progressively expand temporal interaction from local to global scales. This structure is well suited to rPPG signals, which exhibit both short-term continuity and longer-range periodic dependency.}
  \label{fig:intro_comparison}
\end{figure}
To address this limitation, remote photoplethysmography (rPPG) has emerged as a vision-based technique that estimates cardiac activity from facial videos by analyzing subtle appearance variations caused by blood circulation. 

Early rPPG approaches mainly relied on signal processing techniques to recover pulse signals from facial regions. Typical methods employed blind source separation strategies such as independent component analysis (ICA)~\cite{ica} and principal component analysis (PCA)~\cite{pca}, or designed color-space projection models such as CHROM~\cite{chrom} and POS~\cite{pos} to isolate physiological signals from RGB video streams. While these methods exploit physiological priors and optical properties of skin reflectance, they often rely on strong assumptions and remain sensitive to illumination variation, motion, and other real-world disturbances.

Recent advances in deep learning have significantly improved rPPG estimation by learning spatiotemporal representations directly from video sequences~\cite{deepphys, physnet, physformer, rhythmformer, reperio}. Despite this progress, effectively modeling the temporal dynamics of physiological signals remains challenging. Many existing architectures rely on temporal shifting or convolutional operators that aggregate information primarily from neighboring frames, resulting in predominantly local temporal interactions and limited temporal receptive fields~\cite{jamsnet, efficientphys, tscan, bigsmall}. However, cardiac-induced appearance variations evolve over longer temporal spans, and such local interactions restrict the model’s ability to capture long-range temporal dependencies that are important for physiological signal modeling.

To address this limitation, we propose BTS-rPPG, a temporal modeling framework based on Orthogonal Butterfly Temporal Shifting (BTS). Inspired by the butterfly communication pattern in the Fast Fourier Transform (FFT), BTS establishes structured frame interactions through an XOR-based butterfly pairing schedule, progressively expanding the temporal receptive field and enabling efficient information propagation across distant frames.

However, directly exchanging features across frames may introduce redundant information. In many shift-based operators, such as Temporal Channel Shift (TCS)~\cite{tsm} and related variants~\cite{tps}, features are transferred between frames whose representations are often highly correlated, leading to redundant feature propagation and inefficient temporal communication. To alleviate this issue, we further introduce an orthogonal feature transfer (OFT) mechanism that filters the source feature with respect to the target context before temporal shifting, retaining only the orthogonal component for cross-frame transmission. By transmitting only information that is complementary to the target representation, this mechanism reduces redundancy and encourages more effective temporal interaction across frames.

Overall, our contributions can be summarized as follows:

1. We introduce BTS, a structured temporal interaction mechanism that establishes long-range frame communication through an XOR-based butterfly pairing schedule, progressively expanding the temporal receptive field.

2. We propose OFT, which filters the source feature with respect to the target context before temporal shifting, retaining only the orthogonal component for cross-frame transmission and reducing redundant temporal communication.

3. Extensive experiments on multiple benchmark datasets demonstrate that BTS-rPPG effectively captures long-range physiological dynamics and consistently improves rPPG estimation performance.

The rest of this paper is organized as follows. Section~\ref{sec:relatedwork} reviews prior work on rPPG estimation and temporal modeling. Section~\ref{sec:method} introduces the proposed BTS-rPPG framework, including the BTS mechanism and OFT strategy. Section~\ref{sec:experiments} presents the experimental protocol and the main quantitative comparisons. Section~\ref{sec:ablation_study} further analyzes the proposed method through ablation studies and design analyses. Section~\ref{sec:Conclusion} concludes the paper with a discussion of limitations and future work.
\section{Related Works}
\label{sec:relatedwork}

\begin{figure*}[hbt] 
    \centering
    \includegraphics[width=\linewidth]{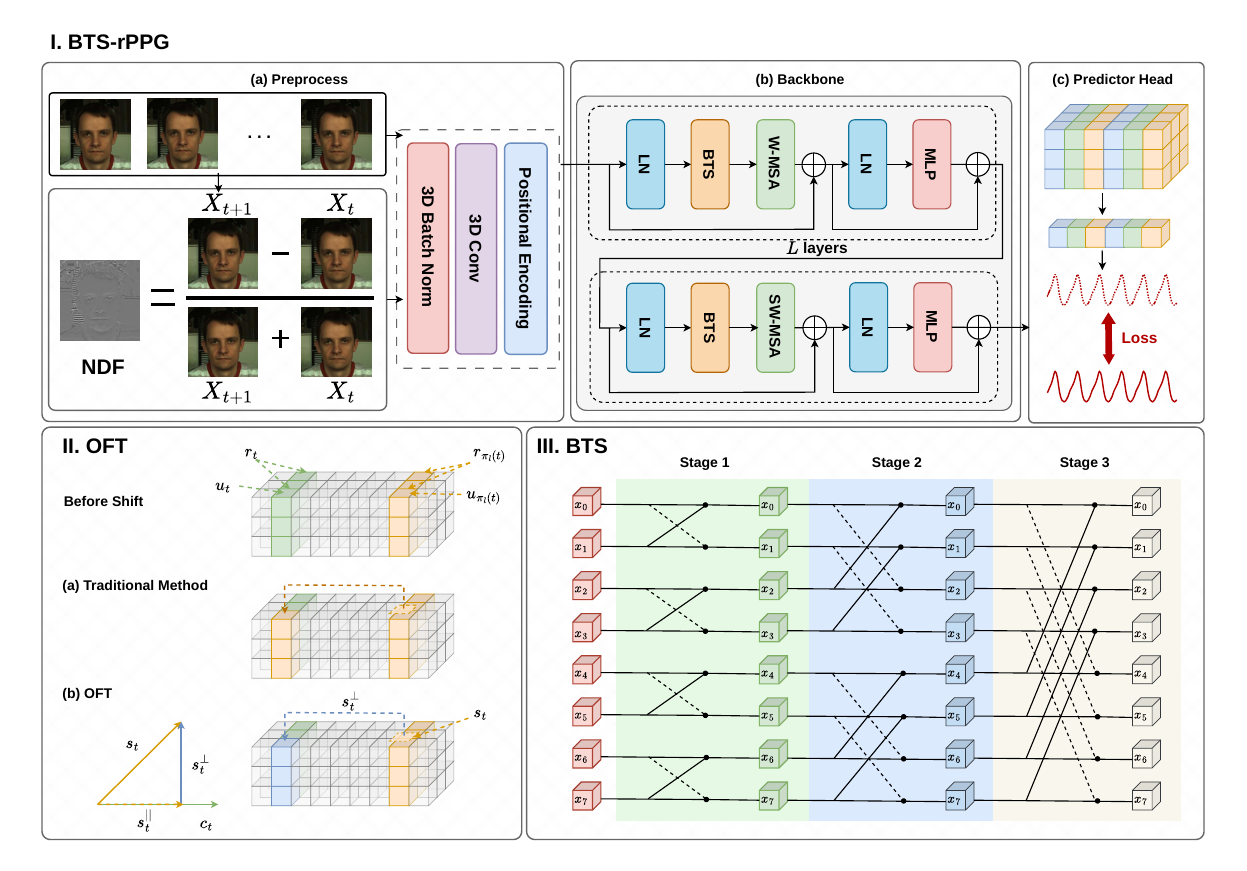} 

    \caption{Overview of the proposed BTS-rPPG. I) The framework comprises (a) an input representation module, (b) a BTS-enhanced Swin Transformer backbone, and (c) a predictor head. RGB frames and their normalized difference frames are fused and embedded into patch-level tokens, which are then processed by Swin Transformer stages equipped with BTS and finally regressed to the target rPPG waveform. II) OFT suppresses redundant transferred information by preserving only the feature component orthogonal to the target context, thereby promoting complementary temporal exchange. III) BTS establishes a butterfly-inspired hierarchical communication pattern, where temporal interactions are progressively expanded across stages to enable efficient propagation from short-range to long-range dependencies.}
    \label{fig:BTS-rPPG} 
\end{figure*}

\subsection{Traditional rPPG Methods}

Early rPPG methods relied on signal processing techniques to recover physiological signals from facial videos. Many approaches exploited color variations caused by blood circulation and applied blind source separation methods such as ICA~\cite{ica} and PCA~\cite{pca} to extract pulse signals from RGB channels. Later works introduced color-space projection models that explicitly modeled the optical properties of skin reflectance. Representative examples include CHROM~\cite{chrom} and POS~\cite{pos}, which project RGB signals onto carefully designed subspaces to enhance pulsatile components while suppressing noise. Although these methods leverage physiological priors and interpretable signal models, their performance is often sensitive to illumination changes, head motion, and other environmental disturbances.

\subsection{Deep Learning-based rPPG}

Recent advances in deep learning have significantly improved rPPG estimation by learning spatiotemporal representations directly from video sequences. Early learning-based approaches employed convolutional neural networks to model spatiotemporal patterns in facial videos~\cite{tscan, efficientphys, deepphys, physnet}. Subsequent works further incorporated temporal attention mechanisms, transformer architectures, or self-supervised learning strategies to improve robustness under challenging conditions~\cite{physformer, rhythmformer, qian2024dual, rs+, rppg-mae, reperio, physdiff, mvrd, physllm}. These methods have demonstrated strong performance across multiple benchmark datasets and have become the dominant paradigm in modern rPPG research.

\subsection{Temporal Modeling for rPPG}

Modeling temporal dynamics plays a critical role in rPPG estimation because physiological signals exhibit structured temporal patterns. Many existing approaches employ temporal convolutions or shift-based operators to exchange information across frames. For example, TCS~\cite{tsm} and Temporal Patch Shift (TPS)~\cite{tps} allow efficient temporal feature propagation without introducing additional parameters. However, such operations typically aggregate information from neighboring frames, resulting in predominantly local temporal interactions. As a consequence, capturing long-range temporal dependencies remains challenging.

To address this limitation, several recent works attempt to enhance temporal modeling through advanced architectures or temporal aggregation strategies~\cite{lsts, reperio}. In contrast to these approaches, our method introduces a structured temporal interaction mechanism based on a butterfly communication pattern. The proposed BTS-rPPG establishes stage-wise frame interactions that progressively expand the temporal receptive field while maintaining efficient computation. Furthermore, our OFT mechanism encourages complementary information exchange between frames, reducing redundancy during temporal communication.

\section{Methodology}
\label{sec:method}

\subsection{Problem Formulation}
\label{subsec:problem_formulation}

Given a facial video clip
\begin{equation}
\mathbf{X}=\{X_t\}_{t=0}^{T-1}, \qquad X_t\in\mathbb{R}^{H\times W\times 3},
\end{equation}
our goal is to estimate the corresponding rPPG waveform
\begin{equation}
\mathbf{y}=[y_0,\dots,y_{T-1}]\in\mathbb{R}^{T}.
\end{equation}
We learn a mapping
\begin{equation}
f_{\theta}:\mathbb{R}^{T\times H\times W\times 3}\rightarrow \mathbb{R}^{T},
\qquad
\hat{\mathbf{y}}=f_{\theta}(\mathbf{X}),
\end{equation}
where \(\hat{\mathbf{y}}=[\hat{y}_1,\dots,\hat{y}_T]\) denotes the predicted rPPG.

As shown in Fig.~\ref{fig:BTS-rPPG}, the proposed BTS-rPPG consists of three components: (a) an input representation module, (b) a BTS-enhanced Swin Transformer backbone, and (c) a predictor head. For notation, let
\begin{equation}
\mathbf{x}^{(\ell)}_{t, n}\in\mathbb{R}^{C_\ell}
\end{equation}
denote the token at frame index \(t\), spatial index \(n\), and stage index \(\ell\), where \(C_\ell\) is the channel dimension at the \(\ell\)-th stage.

\subsection{Input Representation}
\label{subsec:input_representation}

To emphasize subtle temporal appearance changes, we construct a normalized difference frame (NDF) representation:
\begin{equation}
D_t=\frac{X_{t+1}-X_t}{X_{t+1}+X_t},\quad t<T-1,\quad D_{T-1}=0.
\end{equation}
The input representation is then formed by channel-wise concatenating each RGB frame with its corresponding NDF:
\begin{equation}
Z_t=\operatorname{Concat}(X_t,D_t),
\qquad Z_t\in\mathbb{R}^{H\times W\times 6},
\end{equation}
yielding the input video tensor
\begin{equation}
\mathbf{Z}=\{Z_t\}_{t=0}^{T-1}\in\mathbb{R}^{T\times H\times W\times 6}.
\end{equation}
A 3D patch embedding layer projects \(\mathbf{Z}\) into the input features of the Swin Transformer backbone~\ref{subsec:backbone}:
\begin{equation}
\mathbf{F}^{(0)}=\operatorname{PE}\big(\operatorname{Conv3D}(\operatorname{BN}(\mathbf{Z}))\big).
\label{eq:feature_init}
\end{equation}
Here, \(\mathbf{F}^{(0)}\in\mathbb{R}^{T\times N\times C_0}\), where \(N=H'W'\) denotes the number of spatial patch tokens and \(C_0\) is the embedding dimension of each token.

\subsection{Swin Transformer Backbone}
\label{subsec:backbone}

The backbone is composed of \(L\) Swin Transformer stages. Let \(\mathbf{F}^{(\ell-1)}\) and \(\mathbf{F}^{(\ell)}\) denote the input and output of the \(\ell\)-th stage, respectively, where \(\ell\in \overline{1, L}\), and \(\mathbf{F}^{(0)}\) is defined in Eq.~\ref{eq:feature_init}. For the \(\ell\)-th stage, we first obtain the window tokens
\begin{equation}
\mathbf{T}^{(\ell)}
=
\mathcal{R}^{(\ell)}
\!\left(
\operatorname{LN}\!\left(\mathbf{F}^{(\ell-1)}\right)
\right),
\end{equation}
where \(\mathcal{R}^{(\ell)}\) denotes the window-token formation operator at stage \(\ell\). Butterfly temporal communication is then applied to \(\mathbf{T}^{(\ell)}\) before the linear projections for self-attention:
\begin{equation}
\mathbf{S}^{(\ell)}
=
\mathcal{B}^{(\ell)}\!\left(\mathbf{T}^{(\ell)}\right).
\label{eq:applied_butterfly}
\end{equation}
The resulting tokens are subsequently projected to queries, keys, and values:
\begin{equation}
\mathbf{Q}^{(\ell)},\mathbf{K}^{(\ell)},\mathbf{V}^{(\ell)}
=
\Phi_{qkv}\!\left(\mathbf{S}^{(\ell)}\right).
\end{equation}
The attention branch is therefore written as
\begin{equation}
\mathbf{H}^{(\ell)}
=
\mathbf{F}^{(\ell-1)}
+
\left(\mathcal{R}^{(\ell)}\right)^{-1}
\!\left(
\operatorname{MSA}\!\left(\mathbf{Q}^{(\ell)},\mathbf{K}^{(\ell)},\mathbf{V}^{(\ell)}\right)
\right),
\end{equation}
and the stage output is given by
\begin{equation}
\mathbf{F}^{(\ell)}
=
\mathbf{H}^{(\ell)}
+
\operatorname{MLP}\!\left(\operatorname{LN}\!\left(\mathbf{H}^{(\ell)}\right)\right).
\end{equation}
The butterfly operator \(\mathcal{B}^{(\ell)}\) is detailed in Sec.~\ref{subsec:bts}.

\subsection{Orthogonal Butterfly Temporal Shifting}
\label{subsec:bts}

\subsubsection{Butterfly Pairing Schedule}
To progressively enlarge the temporal communication range, we organize frame interactions according to a butterfly schedule. At stage \(\ell\), frame \(t\) communicates with its paired frame
\begin{equation}
\pi_{\ell}(t)=t\oplus 2^{\ell-1},
\qquad \ell=1,\dots,L,
\end{equation}
where \(\oplus\) denotes the bitwise XOR operation. This yields stage-wise temporal offsets
\(1,2,4,\dots,2^{L-1}\), which progressively expand the temporal receptive field across stages.

\subsubsection{Channel Fold Decomposition}

For a token entering the butterfly operator at stage \(\ell\), we express its channel dimension in accordance with the \(H\)-head structure of the Transformer as
\begin{equation}
\mathbf{x}^{(\ell)}_{t}
=
\left[
\mathbf{x}^{(\ell)}_{t,1}
\,;\,
\mathbf{x}^{(\ell)}_{t,2}
\,;\,
\dots
\,;\,
\mathbf{x}^{(\ell)}_{t,H}
\right],
\qquad
\mathbf{x}^{(\ell)}_{t,h}\in\mathbb{R}^{d_\ell},
\end{equation}
where \(d_\ell=C_\ell/H\). For each \(h\), we decompose
\begin{equation}
\mathbf{x}^{(\ell)}_{t,h}
=
\left[
\mathbf{u}^{(\ell)}_{t,h}
\,;\,
\mathbf{r}^{(\ell)}_{t,h}
\right],
\end{equation}
where
\begin{equation}
\mathbf{u}^{(\ell)}_{t,h}\in\mathbb{R}^{C_s},
\qquad
\mathbf{r}^{(\ell)}_{t,h}\in\mathbb{R}^{d_\ell-C_s},
\end{equation}
and \(C_s \ll d_\ell\). Here, \(\mathbf{u}^{(\ell)}_{t,h}\) denotes the transferable fold, while \(\mathbf{r}^{(\ell)}_{t,h}\) remains in the current frame.

For target frame \(t\), the source fold at stage \(\ell\) is taken from its butterfly partner:
\begin{equation}
\mathbf{s}^{(\ell)}_{t,h}
=
\mathbf{u}^{(\ell)}_{\pi_{\ell}(t),h}.
\end{equation}

\subsection{Orthogonal Feature Transfer}
\label{subsec:oft}

Directly transferring \(\mathbf{s}^{(\ell)}_{t,h}\) may introduce redundancy when the source and target frames are highly correlated. To suppress this redundancy, we project the retained target features to the transferable subspace and define the target context as
\begin{equation}
\mathbf{c}^{(\ell)}_{t,h}
=
\mathbf{P}^{(\ell)}_{h}\mathbf{r}^{(\ell)}_{t,h},
\end{equation}
where \(\mathbf{P}^{(\ell)}_{h}\in\mathbb{R}^{C_s\times(d_\ell-C_s)}\) is a learnable projection matrix. The component of the source fold aligned with the target context is
\begin{equation}
\mathbf{s}^{\parallel(\ell)}_{t,h}
=
\frac{
\left\langle
\mathbf{s}^{(\ell)}_{t,h},
\mathbf{c}^{(\ell)}_{t,h}
\right\rangle
}{
\left\|
\mathbf{c}^{(\ell)}_{t,h}
\right\|_2^2+\epsilon
}
\mathbf{c}^{(\ell)}_{t,h},
\end{equation}
and the orthogonal component is
\begin{equation}
\mathbf{s}^{\perp(\ell)}_{t,h}
=
\mathbf{s}^{(\ell)}_{t,h}
-
\mathbf{s}^{\parallel(\ell)}_{t,h}.
\end{equation}
The target token is then updated by replacing its transferable fold with the orthogonalized source fold:
\begin{equation}
\widetilde{\mathbf{x}}^{(\ell)}_{t,h}
=
\left[
\mathbf{s}^{\perp(\ell)}_{t,h}
\,;\,
\mathbf{r}^{(\ell)}_{t,h}
\right].
\end{equation}
Equivalently, the butterfly operator \(\mathcal{B}^{(\ell)}\) can be defined element-wise as
\begin{equation}
\mathcal{B}^{(\ell)}
\left(
\mathbf{x}^{(\ell)}_{t}
\right)
=
\left[
\widetilde{\mathbf{x}}^{(\ell)}_{t,1}
\,;\,
\widetilde{\mathbf{x}}^{(\ell)}_{t,2}
\,;\,
\dots
\,;\,
\widetilde{\mathbf{x}}^{(\ell)}_{t,H}
\right],
\end{equation}
with
\begin{equation}
\widetilde{\mathbf{x}}^{(\ell)}_{t,h}
=
\left[
\operatorname{OFT}
\left(
\mathbf{u}^{(\ell)}_{\pi_{\ell}(t),h},
\mathbf{P}^{(\ell)}_{h}\mathbf{r}^{(\ell)}_{t,h}
\right)
\,;\,
\mathbf{r}^{(\ell)}_{t,h}
\right],
\end{equation}
where
\begin{equation}
\operatorname{OFT}(\mathbf{a},\mathbf{b})
=
\mathbf{a}
-
\frac{
\langle \mathbf{a},\mathbf{b}\rangle
}{
\|\mathbf{b}\|_2^2+\epsilon
}
\mathbf{b}.
\end{equation}

\subsection{Predictor Head and Training Objective}
\label{subsec:head_loss}

After the last backbone stage, we obtain
\begin{equation}
\mathbf{F}^{(L)}\in\mathbb{R}^{T\times N\times C_L}.
\end{equation}
Let \(\mathbf{x}^{(L)}_{t,n}\) denote the \((t,n)\)-th token in \(\mathbf{F}^{(L)}\). Frame-wise spatial aggregation is performed by global average pooling:
\begin{equation}
\mathbf{z}_t
=
\frac{1}{N}
\sum_{n=1}^{N}
\mathbf{x}^{(L)}_{t,n},
\qquad
\mathbf{z}_t\in\mathbb{R}^{C_L}.
\end{equation}
The final rPPG waveform is then predicted by applying a linear projection to each frame-level representation:
\begin{equation}
\hat{\mathbf{y}}
=
\big\|_{t=0}^{T-1}
\operatorname{Linear}(\mathbf{z}_t;1),
\end{equation}
where \(\|\) denotes the concatenation operator along the temporal dimension, and \(\operatorname{Linear}(\cdot;1):\mathbb{R}^{C_L}\rightarrow\mathbb{R}\) maps each \(\mathbf{z}_t\) to a scalar prediction.

The model is optimized using the negative Pearson correlation loss:
\begin{equation}
\mathcal{L}_{\mathrm{NP}}
=
1-
\frac{
\sum_{t=0}^{T-1}
(\hat{y}_t-\bar{\hat{y}})(y_t-\bar{y})
}{
\sqrt{
\sum_{t=0}^{T-1}(\hat{y}_t-\bar{\hat{y}})^2
}
\sqrt{
\sum_{t=0}^{T-1}(y_t-\bar{y})^2
}
}.
\end{equation}
\section{Experiments}
\label{sec:experiments}

\begin{table*}[t]
\centering
\scriptsize
\setlength{\tabcolsep}{2.8pt}
\renewcommand{\arraystretch}{1.05}
\newcommand{\NA}{--}
\resizebox{\linewidth}{!}{%
\begin{tabular}{l*{12}{c}}
\toprule
\multirow{2}{*}{Method} &
\multicolumn{4}{c}{PURE} &
\multicolumn{4}{c}{UBFC-rPPG} &
\multicolumn{4}{c}{MMPD} \\
\cmidrule(lr){2-5}\cmidrule(lr){6-9}\cmidrule(lr){10-13}
& MAE$\downarrow$ & MAPE$\downarrow$ & RMSE$\downarrow$ & $r\uparrow$
& MAE$\downarrow$ & MAPE$\downarrow$ & RMSE$\downarrow$ & $r\uparrow$
& MAE$\downarrow$ & MAPE$\downarrow$ & RMSE$\downarrow$ & $r\uparrow$ \\
\midrule

LGI~\cite{lgi}
& 3.59 & 3.37 & 14.66 & 0.79
& 5.39 & 11.08 & 15.09 & 0.81
& 16.63 & 18.77 & 23.06 & 0.11 \\

CHROM~\cite{chrom}
& 5.39 & 11.08 & 15.09 & 0.81
& 4.06 & 3.34 & 8.83 & 0.89
& 13.73 & 16.95 & 18.88 & 0.15 \\

POS~\cite{pos}
& 0.36 & 0.50 & 0.93 & 1.00
& 4.08 & 3.93 & 7.72 & 0.92
& 15.61 & 18.28 & 21.40 & 0.14 \\

DeepPhys~\cite{deepphys}
& 3.33 & 2.91 & 14.45 & 0.90
& 0.76 & \underline{0.79} & \underline{1.09} & 0.99
& 23.73 & 25.63 & 28.25 & -0.06 \\
 
PhysNet~\cite{physnet}
& 0.93 & 1.40 & 2.08 & 0.99
& 2.25 & 2.37 & 4.81 & 0.94
& \underline{4.81} & \underline{4.84} & \underline{11.83} & 0.60 \\

TS-CAN~\cite{tscan}
& 0.32 & 0.50 & 0.63 & 0.99
& 1.24 & 1.35 & 2.79 & 0.96
& 8.97 & 9.43 & 16.58 & 0.44 \\

PhysFormer~\cite{physformer}
& 0.52 & 0.86 & 1.03 & 0.99
& 2.34 & 2.60 & 5.55 & 0.97
& 13.64 & 14.42 & 19.39 & 0.15 \\

EfficientPhys~\cite{efficientphys}
& 0.55 & 0.71 & 1.34 & 0.99
& \underline{0.73} & 0.83 & 2.53 & 0.97
& 12.79 & 13.48 & 21.12 & 0.24 \\

RhythmFormer~\cite{rhythmformer}
& \underline{0.27} & \underline{0.31} & \underline{0.46} & \underline{1.00}
& 0.81 & 0.80 & 1.12 & \underline{0.99}
& 6.74 & 6.93 & 11.92 & \underline{0.71} \\

\rowcolor{gray!25}
BTS-rPPG (Ours)
& \textbf{0.21} & \textbf{0.25} & \textbf{0.40} & \textbf{1.00}
& \textbf{0.24} & \textbf{0.28} & \textbf{0.36} & \textbf{1.00}
& \textbf{4.05} & \textbf{4.45} & \textbf{9.74} & \textbf{0.75} \\
\bottomrule
\end{tabular}}
\caption{Intra-dataset evaluation results on PURE~\cite{pure}, UBFC-rPPG~\cite{ubfc}, and MMPD~\cite{mmpd}. Lower is better for MAE, MAPE, and RMSE, while higher is better for Pearson correlation $r$. Best results are highlighted in bold and second-best results are underlined.}
\label{tab:intra_dataset_comparison}
\end{table*}

\begin{table*}[t]
\centering
\footnotesize
\setlength{\tabcolsep}{3.5pt}
\renewcommand{\arraystretch}{1.15}
\resizebox{\linewidth}{!}{%
\begin{tabular}{l*{16}{c}}
\toprule
\multirow{2}{*}{Model} &
\multicolumn{4}{c}{PURE $\rightarrow$ UBFC-rPPG} &
\multicolumn{4}{c}{UBFC-rPPG $\rightarrow$ PURE} &
\multicolumn{4}{c}{PURE $\rightarrow$ MMPD} &
\multicolumn{4}{c}{UBFC-rPPG $\rightarrow$ MMPD} \\
\cmidrule(lr){2-5}\cmidrule(lr){6-9}\cmidrule(lr){10-13}\cmidrule(lr){14-17}
& MAE$\downarrow$ & MAPE$\downarrow$ & RMSE$\downarrow$ & $r\uparrow$
& MAE$\downarrow$ & MAPE$\downarrow$ & RMSE$\downarrow$ & $r\uparrow$
& MAE$\downarrow$ & MAPE$\downarrow$ & RMSE$\downarrow$ & $r\uparrow$
& MAE$\downarrow$ & MAPE$\downarrow$ & RMSE$\downarrow$ & $r\uparrow$ \\
\midrule

DeepPhys~\cite{deepphys}
& 1.21 & 1.42 & 2.90 & 0.99
& 8.06 & 13.67 & 19.71 & 0.61
& 16.92 & 18.54 & 24.61 & 0.05
& 17.50 & 19.27 & 25.00 & 0.05 \\

PhysNet~\cite{physnet}
& \underline{0.98} & \underline{1.12} & \underline{2.48} & \underline{0.99}
& \underline{3.69} & \textbf{3.39} & \underline{13.80} & \underline{0.82}
& \underline{13.22} & \underline{14.73} & \underline{19.61} & \underline{0.23}
& \underline{10.24} & \underline{12.46} & \underline{16.54} & \underline{0.29} \\

TSCAN~\cite{tscan}
& 1.30 & 1.50 & 2.87 & 0.99
& 12.92 & 23.92 & 24.36 & 0.47
& 13.94 & 15.14 & 21.61 & 0.20
& 14.01 & 15.48 & 21.04 & 0.24 \\

PhysFormer~\cite{physformer}
& 1.44 & 1.66 & 3.77 & 0.98
& 12.92 & 23.92 & 24.36 & 0.47
& 14.57 & 16.73 & 20.71 & 0.15
& 12.10 & 15.41 & 17.79 & 0.17 \\

EfficientPhys~\cite{efficientphys}
& 2.07 & 2.10 & 6.32 & 0.94
& 5.47 & 5.40 & 17.04 & 0.71
& 14.03 & 15.32 & 21.62 & 0.17
& 13.78 & 15.15 & 22.25 & 0.09 \\

RhythmFormer~\cite{rhythmformer}
& 2.26 & 2.34 & 5.48 & 0.95
& 8.02 & 15.71 & 18.14 & 0.74
& 17.52 & 20.20 & 25.51 & 0.18
& 14.95 & 20.46 & 21.16 & 0.18 \\

\rowcolor{gray!25}
BTS-rPPG (Ours)
& \textbf{0.84} & \textbf{1.08} & \textbf{2.07} & \textbf{1.00}
& \textbf{3.27} & \underline{3.63} & \textbf{8.41} & \textbf{0.94}
& \textbf{10.34} & \textbf{11.83} & \textbf{18.01} & \textbf{0.25}
& \textbf{9.67} & \textbf{13.48} & \textbf{15.39} & \textbf{0.30} \\

\bottomrule
\end{tabular}}
\caption{Cross-dataset test results. PURE$\rightarrow$UBFC-rPPG means the model is trained on PURE and tested on UBFC-rPPG. Lower is better for MAE/MAPE/RMSE, while higher is better for Pearson correlation $r$.}
\label{tab:cross_comparison}
\end{table*}

\subsection{Datasets}

We validate the proposed method on three widely adopted rPPG benchmarks: PURE~\cite{pure}, UBFC-rPPG~\cite{ubfc}, and MMPD~\cite{mmpd}.

\paragraph{PURE} is a controlled rPPG benchmark containing 60 video sequences collected from 10 subjects under six predefined head-motion scenarios, including steady sitting, talking, slow head motion, fast head motion, small head rotation, and medium head rotation. Each video is recorded at a spatial resolution of \(640\times480\) and 30 fps using lossless PNG compression, while the ground-truth PPG signal is synchronously captured by a CMS50E pulse oximeter at 60 Hz. Owing to its clean acquisition setting and explicitly designed motion protocols, PURE is widely used to assess the motion robustness of rPPG methods.

\paragraph{UBFC-rPPG} consists of 42 indoor facial video recordings acquired while participants perform a time-limited mathematical task, which induces natural variations in heart rate. The videos are captured by a webcam at \(640\times480\) resolution and 30 fps, with synchronized ground-truth PPG signals recorded using a CMS50E pulse oximeter. Compared with PURE, UBFC-rPPG presents greater appearance variability due to mixed illumination from natural and artificial light sources, making it a more challenging benchmark.

\paragraph{MMPD} is a large-scale multi-domain rPPG dataset containing approximately 11 hours of facial videos collected from 33 subjects using mobile phone cameras. It explicitly considers variations along three important factors: skin tone (Fitzpatrick types III--VI), illumination condition (high-intensity LED, low-intensity LED, incandescent, and natural light), and activity type (stationary, head rotation, talking, and walking). These diverse acquisition conditions introduce substantial domain shift, making MMPD a challenging benchmark for evaluating the robustness and generalization ability of rPPG models in more realistic scenarios.
\subsection{Implementation Details}
\label{subsec:impl_details}

In this study, facial regions are detected using RetinaFace~\cite{retinaface}. To preserve sufficient facial context, each detected bounding box is enlarged by a factor of \(1.5\) prior to cropping, and the resulting face region is resized to a fixed spatial resolution of \(128\times128\) pixels. During training, random horizontal flipping is applied for data augmentation. A sliding-window protocol is adopted to construct video clips of \(192\) frames with a temporal stride of \(96\). The initial patch embedding dimension is set to \(C_0=192\), and the Swin Transformer~\cite{swintransformer} backbone employs \(8\) attention heads.

For the \(192\)-frame setting, the 6-stage butterfly interaction schedule with temporal offsets \(\{1,2,4,8,16,32\}\) is instantiated as two consecutive butterfly cycles. The first cycle operates on the original temporal ordering, whereas the second applies the same interaction schedule after a one-step cyclic permutation of the sequence, in which the last frame is relocated to the first position. This practical scheduling strategy alleviates boundary-isolated communication arising from the non-power-of-two clip length and facilitates information propagation across temporal partitions.

All models are trained from scratch on an NVIDIA GeForce RTX 3090 GPU using PyTorch 2.0.0 with CUDA 11.8. We employ the Adam optimizer with a batch size of \(4\), an initial learning rate of \(1\times10^{-4}\), and weight decay of \(0.01\). The learning rate is scheduled using OneCycle~\cite{Onecycle} with cosine annealing. Training is conducted for \(10\) epochs on MMPD and \(30\) epochs on PURE and UBFC-rPPG.
\subsection{Main Comparison}
\subsubsection{Intra-dataset Testing} Tab.~\ref{tab:intra_dataset_comparison} presents the intra-dataset evaluation on PURE~\cite{pure}, UBFC-rPPG~\cite{ubfc}, and MMPD~\cite{mmpd}. Overall, BTS-rPPG delivers the most favorable performance across all three benchmarks, consistently outperforming existing methods under both controlled and more challenging conditions. More specifically, relative to the strongest competing method, BTS-rPPG reduces MAE/MAPE/RMSE by 22.2\%/19.4\%/13.0\% on PURE, 67.1\%/64.6\%/67.0\% on UBFC-rPPG, and 15.8\%/8.1\%/17.7\% on MMPD. Its advantage remains evident not only on the relatively clean PURE dataset, but also on UBFC-rPPG and MMPD, where motion, illumination variation, and appearance diversity make remote physiological measurement substantially more difficult. These results indicate that BTS-rPPG is able to recover more reliable physiological cues while remaining resilient to non-physiological interference, thereby achieving stronger estimation accuracy and robustness than prior approaches.
\begin{figure}[t]
  \centering
  \includegraphics[width=0.8\linewidth]{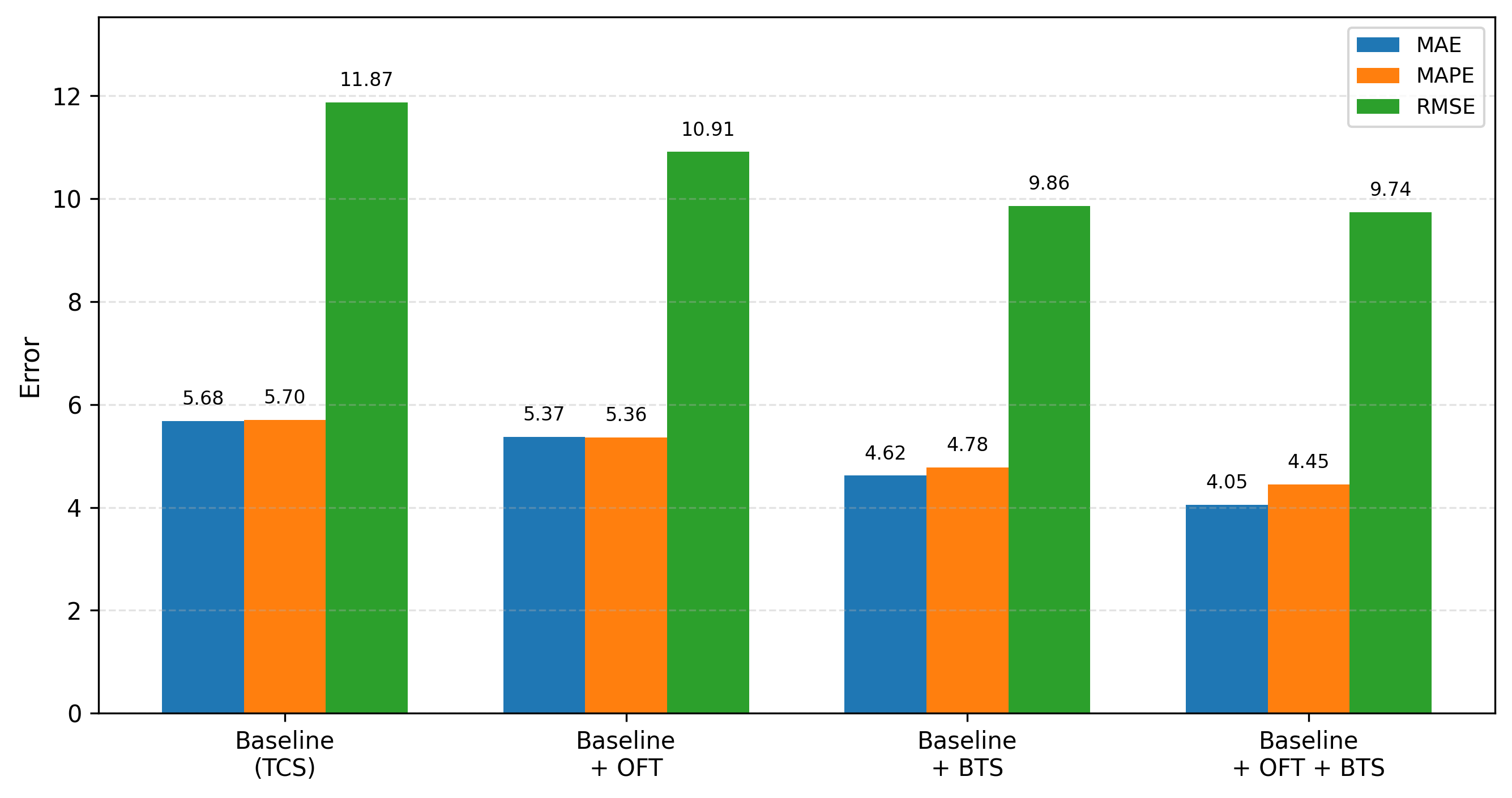}
    \caption{Ablation study on intra-dataset MMPD~\cite{mmpd}.
    This experiment aims to examine how OFT and BTS each contribute to improving the baseline TCS~\cite{tsm} framework.}
    \label{fig:ablation_intra_mmpd}
\end{figure}
\subsubsection{Cross-dataset Testing}
Tab.~\ref{tab:cross_comparison} further evaluates the generalization ability under cross-dataset protocols. Compared with intra-dataset evaluation, cross-dataset testing is considerably more demanding due to the distributional discrepancy across datasets, including differences in recording devices, subject characteristics, motion patterns, and environmental conditions. Nevertheless, BTS-rPPG maintains the best overall transfer performance, suggesting that the proposed model captures more transferable and physiologically meaningful representations rather than relying on dataset-specific visual statistics. This consistent superiority under cross-domain settings highlights the robustness of BTS-rPPG to domain shift and underscores its potential for real-world rPPG deployment.
\subsubsection{Computational Cost and Inference Efficiency}
Table~\ref{tab:efficiency} compares BTS-rPPG with representative baselines in terms of model complexity and inference efficiency. While BTS-rPPG is heavier than lightweight CNN-based methods, it still supports real-time inference on commodity hardware, achieving 2.05\,Kfps on a single NVIDIA T4 GPU. Compared with recent Transformer-based baselines, BTS-rPPG also offers a more favorable efficiency--accuracy trade-off, requiring fewer parameters and MACs than PhysFormer~\cite{physformer}, LSTS~\cite{lsts}, and Reperio-rPPG~\cite{reperio} while delivering higher throughput and lower latency.
\section{Ablation Studies}
\label{sec:ablation_study}

\begin{figure}[t]
  \centering
  \includegraphics[width=\linewidth]{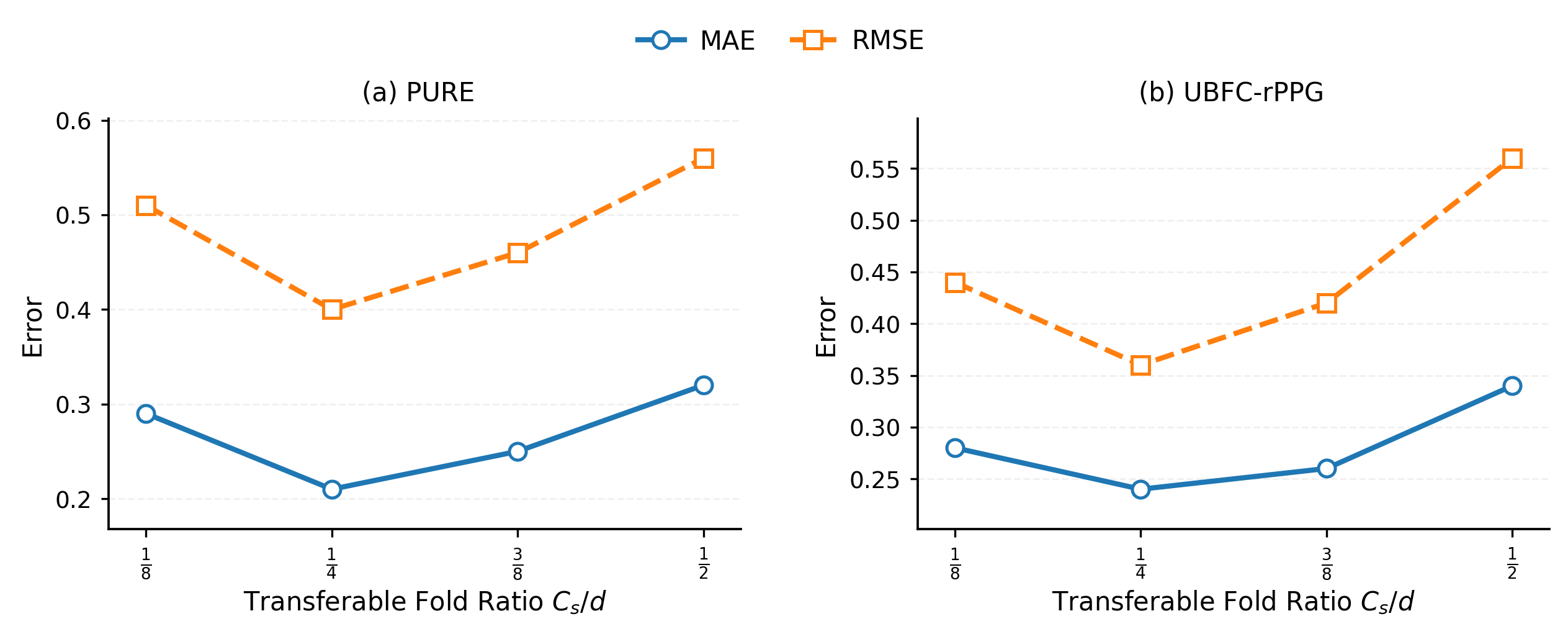}
    \caption{Sensitivity analysis of the OFT fold ratio on (a) PURE~\cite{pure} and (b) UBFC-rPPG~\cite{ubfc} datasets. We report MAE and RMSE, and find that the best performance is achieved at \(C_s/d=1/4\), which provides sufficient cross-frame transfer while retaining adequate target-specific representation.}
    \label{fig:oft_config}
\end{figure}
\begin{table}[t]
\centering
\resizebox{\linewidth}{!}{
\begin{tabular}{lcccc}
\toprule
Method & \#Param (M)$\downarrow$ & MACs (G)$\downarrow$ & Throughput (Kfps)$\uparrow$ & Latency (ms)$\downarrow$ \\
\midrule
MTTS-CAN~\cite{tscan}      & \underline{4.33} & \underline{45.85} & \underline{3.99} & \underline{50.03} \\
EfficientPhys~\cite{efficientphys} & \textbf{2.16}    & \textbf{22.96}    & \textbf{7.21}    & \textbf{27.72} \\
PhysFormer~\cite{physformer}    & 7.38             & 60.72             & 2.27             & 84.50 \\
LSTS~\cite{lsts}          & 6.37             & 71.06             & 1.30             & 147.68 \\
Reperio-rPPG~\cite{reperio}  & 6.50             & 71.02             & 1.47             & 130.18 \\
BTS-rPPG (Ours)      & 5.98             & 69.29             & 2.05             & 98.44 \\
\bottomrule
\end{tabular}
}
\caption{Efficiency comparison of BTS-rPPG against representative baselines. We report the number of parameters, computational cost measured by MACs, throughput, and inference latency under the same benchmarking setting.}
\label{tab:efficiency}
\end{table}
To disentangle the effects of OFT and BTS, we conduct an ablation study on the intra-dataset MMPD~\cite{mmpd} benchmark, as shown in Fig.~\ref{fig:ablation_intra_mmpd}. The results reveal three key findings. First, adding OFT to the baseline consistently reduces MAE, MAPE, and RMSE, indicating that its advantage is not confined to the complete model and can transfer to a simpler backbone. This observation suggests that OFT contributes a broadly applicable improvement mechanism beyond our full framework. Second, BTS delivers stronger gains, highlighting the critical role of hierarchical temporal reasoning in physiological signal estimation. Through the progressive expansion of the temporal receptive field, BTS allows the model to reconcile short-term fluctuations with longer-range rhythmic structure, a property that is particularly important for periodic signals such as rPPG. Third, the joint use of OFT and BTS leads to the best overall performance, demonstrating that the two modules are synergistic: OFT refines feature interaction, while BTS strengthens the representation of temporal dependencies.
\begin{table}[t]
\centering
\small
\setlength{\tabcolsep}{4pt}
\renewcommand{\arraystretch}{1.08}
\resizebox{0.8\columnwidth}{!}{%
\begin{tabular}{@{}l c c c c c c c@{}}
\toprule
\textbf{Variant} & \textbf{$s_1$} & \textbf{$s_2$} & \textbf{$s_3$} & \textbf{$s_4$} & \textbf{$s_5$} & \textbf{$s_6$} & \textbf{RMSE $\downarrow$} \\
\midrule
Local-only         & 1  & 1  & 1 & 1 & 1  & 1  & 14.60 \\
Linear             & 1  & 2  & 3 & 4 & 5  & 6  & 12.18 \\
Reverse Butterfly  & 32 & 16 & 8 & 4 & 2  & 1  & 11.03 \\
Proposed Butterfly & 1  & 2  & 4 & 8 & 16 & 32 & \textbf{9.74} \\
\bottomrule
\end{tabular}%
}
\caption{Design analysis of BTS under different temporal pairing schedules on the intra-dataset MMPD~\cite{mmpd} benchmark.}
\label{tab:bts_schedule}
\end{table}
\subsection{Effect of OFT Configuration}
We study the sensitivity of OFT to the transferable fold width \(C_s\). For each attention head, OFT decomposes the feature into a transferable part \(u_{t,h}^{(\ell)} \in \mathbb{R}^{C_s}\) and a retained part \(r_{t,h}^{(\ell)} \in \mathbb{R}^{d_\ell-C_s}\), where the latter is used to construct the target context for orthogonal filtering. Hence, \(C_s\) controls the balance between cross-frame feature transfer and preservation of target-specific information. 

Fig.~\ref{fig:oft_config} reports MAE and RMSE under four transferable fold ratios, denoted by \(\rho = C_s/d\), with \(\rho \in \{1/8,\,1/4,\,3/8,\,1/2\}\). The optimal performance is obtained at \(\rho = 1/4\) on both PURE~\cite{pure} and UBFC-rPPG~\cite{ubfc}. A smaller ratio (\(\rho = 1/8\)) provides insufficient transferable capacity, whereas larger ratios (\(\rho \in \{3/8,\,1/2\}\)) reduce the retained subspace and thus weaken the target context used by OFT. This leads to less effective orthogonal filtering and degraded performance. Overall, the results indicate that a moderate fold ratio offers the most favorable trade-off between complementary temporal exchange and preservation of frame-specific representation. 
\subsection{Design Analysis of BTS}
\label{sec:design_bts}
We further analyze BTS by varying the temporal pairing schedule across the six interaction stages while keeping the backbone architecture, optimization setting, and training protocol unchanged. Specifically, we compare four temporal schedules: \emph{Local-only}, which repeatedly applies only short-range interaction; \emph{Linear}, which increases the temporal offset in a uniform manner; \emph{Reverse Butterfly}, which begins with large temporal offsets and progressively reduces the interaction range across stages; and the \emph{Proposed Butterfly} schedule, which progressively expands the interaction range from local to long-range dependencies. As reported in Tab.~\ref{tab:bts_schedule}, the proposed butterfly schedule yields the lowest RMSE of 9.74, outperforming all alternative temporal schedules, including Local-only (14.60), Linear (12.18), and Reverse Butterfly (11.03). These results indicate that the benefit of BTS cannot be attributed solely to the inclusion of long-range temporal interaction. Rather, its effectiveness appears to depend critically on how temporal dependency is introduced across network depth. Specifically, the butterfly schedule enforces a coarse-to-fine progression in temporal receptive field expansion, allowing early stages to consolidate short-range physiological dynamics before later stages integrate broader rhythmic structure. By contrast, the Local-only variant lacks sufficient temporal coverage, the Linear schedule provides weaker multiscale separation across stages, and the Reverse Butterfly strategy introduces large temporal offsets prematurely, which may destabilize low-level physiological cues before they are well refined. This comparison suggests that the ordering of temporal interaction is a key design factor, and that a progressively expanded interaction pattern is better aligned with the hierarchical nature of physiological signal formation.
\section{Conclusion}
\label{sec:Conclusion}
In this paper, we presented BTS-rPPG, a temporal modeling framework for rPPG based on BTS. By introducing an FFT-inspired butterfly pairing schedule, BTS enables structured long-range temporal interactions and efficient information propagation across exponentially increasing temporal distances. In addition, OFT filters the source feature with respect to the target context before temporal shifting, retaining only the orthogonal component for cross-frame transmission and thereby reducing redundant feature propagation. Extensive experiments on multiple benchmark datasets demonstrate that BTS-rPPG improves the modeling of long-range physiological dynamics and consistently outperforms existing temporal modeling strategies for rPPG estimation. Despite its strong empirical performance, BTS still lacks a deeper theoretical understanding. Although the butterfly pairing schedule is motivated by FFT-inspired multistage interaction, why this particular stage-wise temporal communication pattern works well for rPPG modeling remains insufficiently explained. In future work, we will further examine the relationship between the temporal interactions induced by BTS and the physiological spectral structure of rPPG signals.
\section*{Acknowledgment}

This work was supported by VinUniversity’s Seed Grant Program under project VUNI.2425.EME.005 and the National Foundation for Science and Technology Development (NAFOSTED) through
Project IZVSZ2-229539 (2025–2027). The authors would like to sincerely thank the Vietnam Young Talent Support Fund, Tan Hiep Phat Trading – Service Co., Ltd., and the Ben Dam Me Award Fund for their valuable support and encouragement of this work.  
\label{sec:acknowledgment}
{
    \small
    \bibliographystyle{ieeenat_fullname}
    \bibliography{main}
}

\end{document}